\documentclass[11pt]{article}
\usepackage[preprint]{acl}
\usepackage{times,latexsym}
\usepackage[T1]{fontenc}
\usepackage[utf8]{inputenc}
\usepackage{microtype}
\usepackage{booktabs,multirow,xcolor,colortbl,array,tabularx}
\definecolor{bfclbest}{RGB}{251,234,191}
\definecolor{bfclsecond}{RGB}{208,229,207}
\definecolor{bfclgray}{RGB}{240,240,240}
\providecommand{\best}[1]{\cellcolor{bfclbest}\textbf{#1}}
\providecommand{\secondbest}[1]{\cellcolor{bfclsecond}\underline{#1}}
\providecommand{\bfclmetric}[1]{{\fontsize{8.5}{10}\selectfont #1}}
\IfFileExists{inconsolata.sty}{\usepackage{inconsolata}}{}
\usepackage{graphicx,amsmath,amssymb,booktabs,multirow,xcolor,placeins}
\usepackage{hyperref}

\title{Where to Start, How to Move: \\
Analyzing Curricula under GRPO for Long-Horizon LLM Agents}
\author{
  Minchan Kim$^{1}$\thanks{Equal contribution.} \quad
  Sungho Bae$^{1}$\footnotemark[1] \quad
  Youngjin Kim$^{2}$ \quad Hyunwoo Park$^{1}$ \\
  {\small $^{1}$Graduate School of Data Science, Seoul National University} \\
  {\small $^{2}$Department of Computer Science and Engineering, Konkuk University} \\
  {\small\texttt{\{mmm5373,sunghobae,hyunwoopark\}@snu.ac.kr}
  \quad\texttt{ab960119@konkuk.ac.kr}}
}
\hypersetup{
  pdftitle={Where to Start, How to Move: Analyzing Curricula under GRPO for Long-Horizon LLM Agents},
  pdfauthor={Minchan Kim, Sungho Bae, Youngjin Kim, Hyunwoo Park}
}
\begin{document}
\maketitle
\begin{abstract}
Group Relative Policy Optimization (GRPO) provides no reward-relative learning signal when all rollouts receive identical rewards. In long-horizon tasks that require multiple interactions, rewards are given only upon successful task completion, making all-fail groups common early in training and limiting opportunities to practice later steps. Guide Assisted GRPO replays a successful reference prefix before the learner continues. We study how initial positions, handoff units, and movement rules affect unassisted task completion. Across three model sizes on BFCL, starting near completion is more effective under a fixed turn-level threshold rule. Finer handoffs alone do not consistently help, but outcome-informed tool-call selection outperforms GRPO on trained and untrained task categories. Guided starts also improve AppWorld success on both evaluation splits. These findings highlight where learning starts and how starting positions move as key curriculum choices.
\end{abstract}

\section{Introduction}
\label{sec:intro}

Group Relative Policy Optimization
(GRPO; \citealp{shao2024deepseekmath}) is widely used for
post-training large language models (LLMs), including interactive
agents \citep{feng2025gigpo}.
It estimates advantages by comparing rewards within groups of rollouts
for the same task, avoiding the cost of training a separate value model.
For agents, these rollouts comprise tool calls and observations
\citep{yao2023react,trivedi2024appworld}, with rewards often
provided only at termination \citep{chen2025loop}.
In long-horizon tasks, early failures can yield all-fail groups and limit practice on later steps \citep{wang2026agentg2}.
Groups with identical outcomes have zero reward-relative advantages.
DAPO replaces these groups \citep{yu2025dapo}, but retries
under the same policy can again all fail.

\begin{figure}[t]
\centering
\captionsetup{skip=4pt}
\includegraphics[width=1\columnwidth]{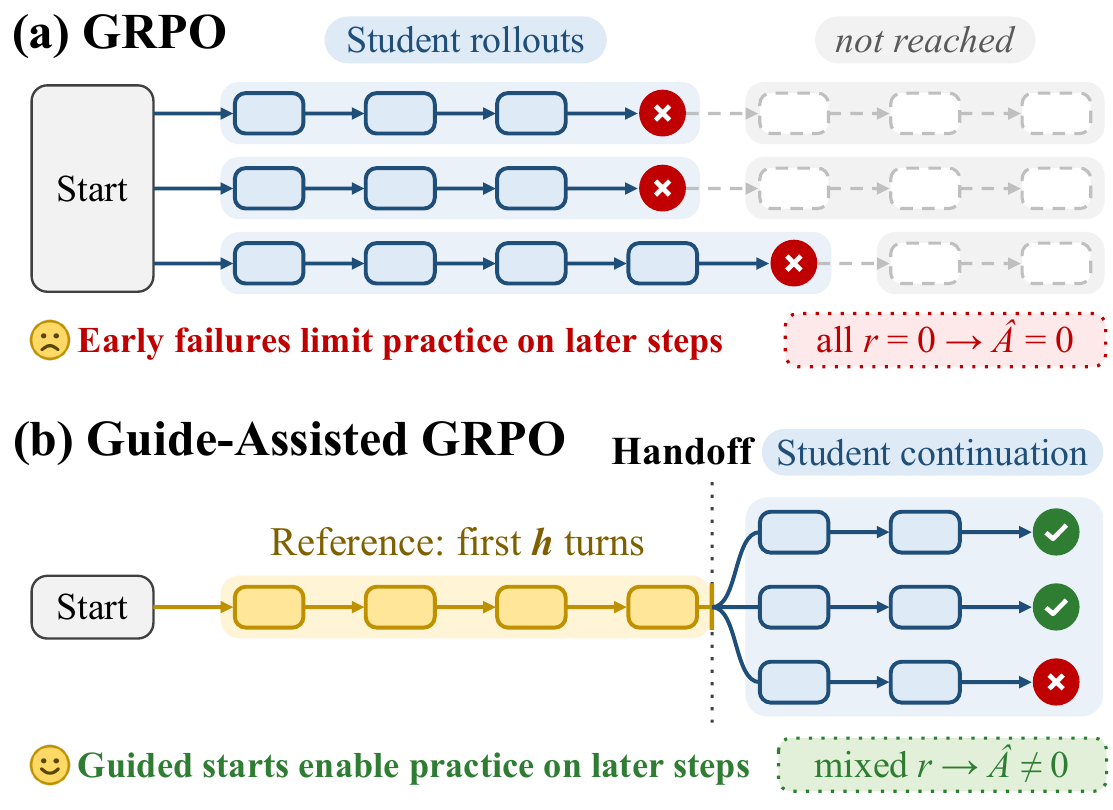}
\caption{GRPO with original and guide-assisted starts.}
\label{fig:guided-start-motivation}
\end{figure}

Starting-state curricula have been explored in general RL through JSRL
\citep{uchendu2023jumpstart}, language-model reasoning through R$^3$
\citep{xi2024reverse}, and interactive agent training through
Agent-G$^2$ \citep{wang2026agentg2}. Drawing on this line of work,
we use \emph{guide-assisted starts}, in which initial steps from
a reference trajectory are executed before the learner continues the
task (Figure~\ref{fig:guided-start-motivation}).

Starting later may make success accessible, but can also remove
reward variation if every continuation succeeds.
As the policy learns, a useful starting point may need to move.
We study how the initial position, handoff unit, and movement rule
relate to completing full tasks without assistance under GRPO.

BFCL experiments span 1.7B, 4B, and 8B models.
Since agents must handle interaction conditions absent from training,
we examine performance both within the trained BFCL category and
across untrained categories.
Separately trained AppWorld agents test whether guided starts also
help with interactive coding that changes application state.

Our BFCL comparisons favor starting near task completion under the same
turn-level movement rule. Finer handoff configurations do not
consistently improve performance; their benefits vary with how training
moves between starts.
Adapting starts using past outcomes at the tool-call level improves
over GRPO on the trained category and, on average, across untrained
categories at all three model sizes. Guided training also improves
AppWorld success rates on both evaluation splits. Together, these
findings support starting-state curricula across tool-use and
interactive coding tasks, with their effectiveness shaped by the
combination of handoff unit and movement rule.
Appendix~\ref{app:related} provides additional related work.
\section{Study Design}
\label{sec:analysis}

\paragraph{Guide-assisted starts.}
A successful reference trajectory serves as a \emph{guide}.
Each rollout resets the environment and replays a guide prefix up to
handoff $h$ before the learner takes control.
Replay changes the environment state and supplies the executed actions
and observations as dialogue history.
The learner completes the episode without access to the unexecuted
guide suffix.

\paragraph{Handoff unit.}
\textbf{Turn} places handoffs at BFCL user-turn boundaries, such as
before a payment request requiring a recipient lookup followed by
a transfer.
\textbf{Call} additionally permits validated boundaries between tool
calls, allowing handoff after the lookup.

\paragraph{Initialization.}
\textbf{Start} initializes at $h=0$, with no replay.
For both Turn and Call, \textbf{End} initializes immediately before
the guide's final user turn.
Both initializations allow subsequent movement in either direction.

\paragraph{Movement.}
Each group contains 16 continuations of the same task from the same
handoff. After each group, \textbf{Threshold} moves earlier after
12--16 successes, later after 0--1, and otherwise stays.
Step sizes depend on the unit (Appendix~\ref{app:selector}).
\textbf{Uniform} samples uniformly from eligible handoffs without
using outcomes.
\textbf{Bandit} tracks outcomes by task and position and uses Thompson
sampling to favor handoffs likely to produce both successes and failures.
Uniform and Bandit can stay or jump in either direction.

\paragraph{Training and evaluation.}
Policy loss applies only to learner-generated tokens.
A binary terminal task reward evaluates the whole episode, including
the guide prefix and learner continuation.
Failure may occur well after handoff, so it does not identify an
error at that position.
All-fail/all-pass groups are filtered from reward-relative updates,
but count toward rollout budgets and inform outcome-based movement.
Evaluation uses $h=0$, requiring the learner to complete the full
task without assistance.

\paragraph{Comparison scope.}
Start/End comparisons hold the turn-level threshold rule fixed.
Call-level comparisons examine movement rules.
Turn/Call comparisons also differ in step sizes or outcome-history
settings, so they compare configurations rather than isolate the
unit's effect.

\paragraph{Protocol and measurements.}
\label{sec:setup}
BFCL experiments use Qwen3-1.7B, Qwen3-4B-Instruct-2507, and
Qwen3-8B \citep{qwen3technicalreport} in non-thinking mode,
trained on Base only
\citep{patil2025bfcl,lu2026dontjustfinetuneagent}.
AppWorld uses separately trained Qwen3-8B agents with handoffs between
code turns \citep{trivedi2024appworld}.
Avg@3 averages three evaluation outcomes per task; Best@3 records
whether any succeeds. Macro Avg. averages the untrained Missing
Function, Missing Parameter, and Long Context categories.
Evaluation repeats do not measure training-seed variability.

\section{Empirical Analysis}
\label{sec:experiments}

\subsection{Does Assisted Training Improve Independent Execution?}
\label{sec:performance}
\begin{table}[h]
\centering\footnotesize
\caption{BFCL Base results (\%). Start and End use turn-level threshold
curricula. {\setlength{\fboxsep}{0.5pt}%
\protect\colorbox{bfclbest}{\strut\textbf{Bold}} and
\protect\colorbox{bfclsecond}{\strut\underline{underlining}}}
mark the best and second-best per model, excluding Initial.}
\label{tab:bfcl-initialization}
\setlength{\tabcolsep}{0pt}
\begin{tabularx}{\columnwidth}{@{}>{\raggedright\arraybackslash}p{43pt}*{6}{>{\centering\arraybackslash}X}@{}}
\toprule
\textbf{Backbone} & \multicolumn{2}{c}{\textbf{1.7B}} & \multicolumn{2}{c}{\textbf{4B}} & \multicolumn{2}{c}{\textbf{8B}}\\
\cmidrule(lr){2-3}\cmidrule(lr){4-5}\cmidrule(lr){6-7}
Method & \bfclmetric{Avg@3} & \bfclmetric{Best@3} & \bfclmetric{Avg@3} & \bfclmetric{Best@3} & \bfclmetric{Avg@3} & \bfclmetric{Best@3}\\
\midrule
Initial & 16.3 & 25.0 & 23.7 & 25.0 & 38.0 & 38.0\\
GRPO & \secondbest{22.3} & 34.0 & \secondbest{54.3} & \best{59.0} & 52.0 & 54.0\\
DAPO & 19.7 & 29.0 & 42.0 & 44.0 & \secondbest{58.7} & \best{67.0}\\
\midrule
\multicolumn{7}{@{}l}{\textit{Turn--Threshold}}\\
Start & 22.0 & \secondbest{35.0} & 53.7 & 56.0 & 55.0 & 56.0\\
End & \best{24.3} & \best{36.0} & \best{55.0} & \secondbest{57.0} & \best{59.0} & \secondbest{61.0}\\
\bottomrule
\end{tabularx}
\end{table}

\paragraph{Refilling groups at the original starting state.}
GRPO provides the unassisted baseline, with all rollouts beginning
at $h=0$. DAPO also starts at $h=0$, but samples additional
rollouts to replace groups with identical rewards.
At the same generated-rollout budget, DAPO trails GRPO in
Avg@3 at 1.7B and 4B and exceeds it at 8B
(Table~\ref{tab:bfcl-initialization}).
Refilling usable groups therefore does not consistently improve
independent task completion in these comparisons.

\paragraph{The benefit of assistance depends on initialization.}
Turn-Threshold with Start initialization yields similar
Base Avg@3 to GRPO at 1.7B and 4B and a higher score at 8B.
With End initialization, Avg@3 is higher than GRPO
at all three sizes, although the gain is modest at 4B.
The largest gain occurs at 8B, reaching 7.0 percentage points.
End initialization also yields higher Avg@3 and Best@3
than Start initialization throughout.
These comparisons favor beginning training with short
continuations near the guide's end.

Replaying a longer guide prefix reduces the portion of
the rollout generated by the learner.
In the Qwen3-8B comparison, Turn-Threshold with End
initialization uses approximately 8\% less training GPU time
than GRPO under the same training setup
(Figure~\ref{fig:signal_access} right).

We next fix End initialization and examine how the handoff unit and movement rule affect performance.

\subsection{Finer Units or Better Movement Rules?}
\label{sec:granularity}
\begin{table}[h]
\centering\footnotesize
\caption{BFCL Base results (\%) by handoff granularity and selection.
All conditions use End initialization.}
\label{tab:bfcl-granularity-selection}
\setlength{\tabcolsep}{0pt}
\begin{tabularx}{\columnwidth}{@{}>{\raggedright\arraybackslash}p{43pt}*{6}{>{\centering\arraybackslash}X}@{}}
\toprule
\textbf{Backbone} & \multicolumn{2}{c}{\textbf{1.7B}} & \multicolumn{2}{c}{\textbf{4B}} & \multicolumn{2}{c}{\textbf{8B}}\\
\cmidrule(lr){2-3}\cmidrule(lr){4-5}\cmidrule(lr){6-7}
Method & \bfclmetric{Avg@3} & \bfclmetric{Best@3} & \bfclmetric{Avg@3} & \bfclmetric{Best@3} & \bfclmetric{Avg@3} & \bfclmetric{Best@3}\\
\midrule
\multicolumn{7}{@{}l}{\textit{Turn}}\\
Threshold & 24.3 & 36.0 & 55.0 & 57.0 & \secondbest{59.0} & 61.0\\
Bandit & 23.0 & 36.0 & \secondbest{57.7} & \secondbest{61.0} & 53.0 & 69.0\\
\midrule
\multicolumn{7}{@{}l}{\textit{Call}}\\
Threshold & 23.3 & 32.0 & 54.3 & 58.0 & 56.0 & 70.0\\
Uniform & \secondbest{29.0} & \best{41.0} & 56.7 & 60.0 & 56.0 & \best{74.0}\\
Bandit & \best{30.0} & \secondbest{40.0} & \best{60.0} & \best{63.0} & \best{60.7} & \secondbest{73.0}\\
\bottomrule
\end{tabularx}
\end{table}

\paragraph{A finer unit is not sufficient.}
Across 100 BFCL training tasks, call-level handoffs provide 557 candidates compared with 366 at the turn level, an increase of about 52\%.
However, Call-Threshold has lower Base Avg@3 than Turn-Threshold
at all three model sizes
(Table~\ref{tab:bfcl-granularity-selection}).
These configuration-level comparisons show that access to more
closely spaced handoffs does not automatically yield better
independent execution.

\paragraph{Bandit gains differ between handoff units.}
Bandit uses outcome histories and stochastic selection to move
among handoffs at either the turn or call level.
Turn-Bandit shows no consistent Base Avg@3 advantage over
Turn-Threshold across model sizes.
In contrast, Call-Bandit exceeds Call-Threshold at all three sizes.
The observed benefit of Bandit therefore depends on the handoff
configuration in which it operates.

To examine whether the call-level gains can be explained by
random movement alone, we compare Call-Bandit with Call-Uniform.
This control samples from the same eligible handoffs without
using rollout outcomes.
Call-Uniform improves or matches Call-Threshold in Base Avg@3,
but Call-Bandit exceeds it by 1.0, 3.3, and 4.7 percentage
points at 1.7B, 4B, and 8B, respectively.
These call-level comparisons suggest that outcome-informed
movement provides benefits beyond uniform random movement.
Appendix Figure~\ref{fig:start_position} details the call-level
starting positions.

\begin{figure}[h]
\centering
\captionsetup{skip=4pt}
\includegraphics[width=0.9\columnwidth]{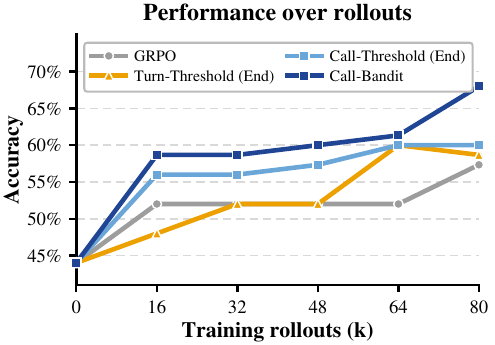}
\caption{BFCL validation accuracy (\%) over training rollouts
with Qwen3-8B.}
\label{fig:val-acc}
\end{figure}

\paragraph{Call-Bandit makes further gains during training.}
In the runs shown in Figure~\ref{fig:val-acc}, Call-Threshold
and Call-Bandit improve rapidly during the first 16k rollouts.
Call-Bandit subsequently makes further gains, while
Call-Threshold improves more gradually and Turn-Threshold
fluctuates toward the end.
Call-Bandit reaches the highest final validation accuracy
among the plotted runs, consistent with its higher Base Avg@3.

\begin{figure*}[t]
\centering
\begin{minipage}[t]{0.705\textwidth}
    \vspace{0pt}
    \centering
    \includegraphics[width=\linewidth]{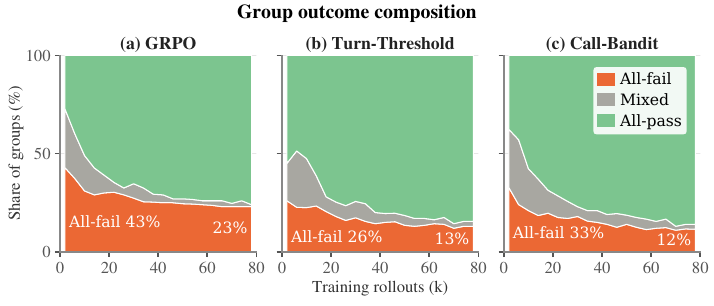}
\end{minipage}%
\begin{minipage}[t]{0.295\textwidth}
    \vspace{0pt}
    \centering
    \includegraphics[width=\linewidth]{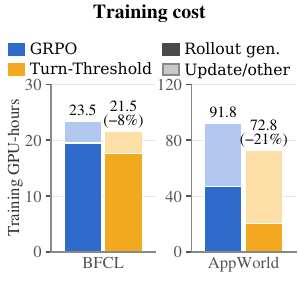}
\end{minipage}

\captionsetup{
    skip=4pt,
    margin=0pt,
    justification=raggedright,
    singlelinecheck=false
}
\caption{(Left) Training-group composition over generated rollouts.
(Right) Training GPU time on BFCL and AppWorld.
All results use Qwen3-8B. Turn-Threshold starts from the last handoff position (End initialization).}
\label{fig:signal_access}
\end{figure*}

\paragraph{Both guided configurations sustain lower all-fail rates.}
Turn-Threshold and Call-Bandit both improve independent task
completion over GRPO at 8B and share a common training pattern:
fewer all-fail groups from early to late training
(Figure~\ref{fig:signal_access} left).
At the first recorded point, all-fail groups account for 43\%
under GRPO, compared with 26\% under Turn-Threshold and 33\%
under Call-Bandit.
By the end of training, these shares are 23\%, 13\%, and 12\%,
respectively.
Thus, the lower incidence of groups without a successful
continuation persists beyond the initial stages of guided training.

We next examine whether the gains observed on Base extend
to untrained BFCL categories.

\subsection{Do Gains Extend to Unseen Interactions?}
\label{sec:cross_category}

Agents must handle missing arguments or functions and information
buried in long observations. BFCL's Missing Parameter, Missing
Function, and Long Context categories test these conditions
\citep{patil2025bfcl}. 

\begin{table}[h]
\centering\footnotesize
\caption{BFCL cross-category generalization (Macro Avg., \%). Scores average Missing Function, Missing Parameter, and Long Context, excluding Base. Training uses only Base tasks. Curricula use End initialization.}
\label{tab:bfcl-ood-compact}
\setlength{\tabcolsep}{0pt}
\begin{tabularx}{\columnwidth}{@{}>{\raggedright\arraybackslash}p{43pt}*{6}{>{\centering\arraybackslash}X}@{}}
\toprule
\textbf{Backbone} & \multicolumn{2}{c}{\textbf{1.7B}} & \multicolumn{2}{c}{\textbf{4B}} & \multicolumn{2}{c}{\textbf{8B}}\\
\cmidrule(lr){2-3}\cmidrule(lr){4-5}\cmidrule(lr){6-7}
Method & \bfclmetric{Avg@3} & \bfclmetric{Best@3} & \bfclmetric{Avg@3} & \bfclmetric{Best@3} & \bfclmetric{Avg@3} & \bfclmetric{Best@3}\\
\midrule
Initial & 11.8 & 19.3 & 19.7 & 21.7 & 41.0 & 56.3\\
GRPO & 15.7 & 26.7 & 40.3 & 43.0 & 45.7 & 58.7\\
DAPO & 14.7 & 24.0 & 32.5 & 36.0 & 45.4 & 56.7\\
\midrule
\multicolumn{7}{@{}l}{\textit{Turn}}\\
Threshold & 14.8 & 24.3 & 34.7 & 38.7 & 46.1 & \secondbest{61.0}\\
Bandit & 17.2 & 28.0 & 35.1 & 38.3 & 46.2 & \secondbest{61.0}\\
\midrule
\multicolumn{7}{@{}l}{\textit{Call}}\\
Threshold & \secondbest{19.0} & \best{30.7} & \secondbest{41.2} & \secondbest{45.3} & \secondbest{46.8} & 60.3\\
Uniform & \secondbest{19.0} & \secondbest{29.3} & 36.9 & 40.7 & 45.8 & 59.0\\
Bandit & \best{19.8} & \best{30.7} & \best{42.7} & \best{45.7} & \best{49.7} & \best{64.7}\\
\bottomrule
\end{tabularx}
\end{table}

\paragraph{The Base ordering does not always persist.}
Call-Threshold exceeds Turn-Threshold in Macro Avg@3
at all three sizes, reversing their Base ordering
(Table~\ref{tab:bfcl-ood-compact}).
Uniform's Base gains do not carry over:
it matches Call-Threshold at 1.7B and trails it at 4B and 8B.
The relative performance of handoff configurations
therefore depends on the evaluation conditions.

\paragraph{Call-Bandit leads across model sizes.}
Call-Bandit achieves the highest Macro Avg@3 at all three sizes,
with gains of 2.4 to 4.1 percentage points over GRPO.
It also exceeds Uniform throughout.
Its higher average success thus extends to the aggregate
of the three untrained categories.

\subsection{Do Gains Extend to Interactive Coding?}
\label{sec:appworld_results}
\label{sec:capability}
\begingroup
\setlength{\intextsep}{8pt plus 2pt minus 2pt}
\begin{table}[h]
\centering\footnotesize
\caption{AppWorld results (success rate, \%) on Test Normal and Test Challenge with Qwen3-8B.}
\label{tab:appworld-main}
\setlength{\tabcolsep}{0pt}
\begin{tabularx}{\columnwidth}{
@{}>{\raggedright\arraybackslash}p{0.41\columnwidth}
*{4}{>{\centering\arraybackslash}X}@{}
}
\toprule
\multirow{2}{*}{Method}
& \multicolumn{2}{c}{Test Normal}
& \multicolumn{2}{c}{Test Challenge}\\
\cmidrule(lr){2-3}\cmidrule(lr){4-5}
& \bfclmetric{Avg@3} & \bfclmetric{Best@3}
& \bfclmetric{Avg@3} & \bfclmetric{Best@3}\\
\midrule
Initial & 5.8 & 10.7 & 1.7 & 2.4\\
GRPO & 43.1 & 54.8 & 20.9 & 28.5\\
Call-Threshold & \best{47.2} & \best{60.7}
& \best{22.3} & \best{31.4}\\
\bottomrule
\end{tabularx}
\end{table}
\endgroup
AppWorld agents interact with the environment by executing
Python code blocks \citep{trivedi2024appworld}, typically over more turns than in BFCL. We refer to handoffs at these code-execution boundaries as Call-level handoffs.

Call-Threshold exceeds GRPO on both Test Normal and Test Challenge
(Table~\ref{tab:appworld-main}).
Its Avg@3 is higher by 4.1 and 1.4 percentage points,
respectively, while Best@3 increases by 5.9 and 2.9 points.
As in BFCL, guide-assisted starts also offer a training-cost
benefit: Call-Threshold uses approximately 21\% less training
GPU time than GRPO (Figure~\ref{fig:signal_access}, right).
Starting near the guide's end can shorten the continuation
the student must generate.

\section{Conclusion}
\label{sec:conclusion}

Starting-state curricula can improve independent task completion
under GRPO, with benefits depending on initialization and movement.
On BFCL, End initialization outperforms Start under
the same turn-level Threshold rule, while finer handoffs alone
do not consistently help. Call-Bandit achieves the highest
Base Avg@3 and cross-category Macro Avg@3 at all three model sizes.
Guide-assisted training also improves AppWorld performance
on both evaluation splits. These findings highlight where
training begins and how it moves between handoffs as key
choices for improving independent execution.
\FloatBarrier
\section*{Limitations}
\label{sec:limitations}

We evaluate guide-assisted training across two benchmarks with
 different interaction structures, BFCL and AppWorld, and include
 three Qwen3 model sizes in the BFCL experiments. This provides
 evidence beyond a single task setting, although further evaluation
 is needed to establish whether the observed trends extend to
 other model families and environments.

To make comparisons more controlled, we hold the turn-level
 movement rule fixed when comparing initial positions and compare
 movement rules within the call-level setting. However, Turn/Call
 configurations also differ in movement step sizes or outcome-history
 settings, so their differences cannot be attributed solely to
 handoff granularity. We also evaluate each task three times to
 account for variation in generated outcomes. These repeated
 evaluations do not capture variability across independent training
 runs, which requires additional training repetitions to assess.

Our experiments use successful reference trajectories to provide
 reproducible starting states. In settings without such trajectories,
 a possible alternative is to generate candidate guides using another
 model and verify their success in the environment. The additional
 collection cost and the influence of guide quality would need
 to be evaluated before extending this approach to those settings.
\section*{Ethical Considerations}
\label{sec:ethics}

This study uses publicly released BFCL and AppWorld benchmarks
and Qwen3 models to analyze agent training.
The BFCL dataset and the Qwen3 checkpoints used here are
released under the Apache License 2.0.
AppWorld's public portion uses the same license; its protected
portion has an additional condition requiring that portion
and its derivatives to be encrypted when publicly
redistributed.\footnote{License information:
\href{https://huggingface.co/datasets/gorilla-llm/Berkeley-Function-Calling-Leaderboard}{BFCL};
\href{https://huggingface.co/Qwen/Qwen3-1.7B/blob/main/LICENSE}{Qwen3-1.7B},
\href{https://huggingface.co/Qwen/Qwen3-4B-Instruct-2507/blob/main/LICENSE}{Qwen3-4B-Instruct-2507},
\href{https://huggingface.co/Qwen/Qwen3-8B/blob/main/LICENSE}{Qwen3-8B};
\href{https://github.com/StonyBrookNLP/appworld\#-license}{AppWorld}.}

Our evaluation measures task completion in benchmark environments
\citep{patil2025bfcl,trivedi2024appworld}; it does not establish
safe deployment with access to real accounts or sensitive information.
Incorrect tool calls or generated code could modify unintended
data or expose private information.
Such deployments require separate safety evaluation and safeguards,
such as restricted permissions, isolated execution, and human
approval for consequential actions.
We do not evaluate the effectiveness of these safeguards. AI assistants were used to organize code and refine the manuscript.

\section*{Acknowledgments}
This work was supported by the ``National AI Project'' of the Ministry of Science and ICT and the AIIS.

\bibliography{references}
\appendix
\IfFileExists{sections/07_appendix.tex}{%
  \section{Additional Related Work}
\label{app:related}

\paragraph{Starting-state curricula.}
JSRL, R$^3$, and Agent-G$^2$ provide different mechanisms for
controlling how much of an episode or solution the learner completes
\citep{uchendu2023jumpstart,xi2024reverse,wang2026agentg2}.
Two related approaches further distinguish access to intermediate
states from the objective optimized afterward.
BREAD combines adaptive expert prefixes with branched rollouts
when self-generated reasoning fails \citep{zhang2025bread}.
PivotRL trains on selected intermediate turns using local on-policy
rollouts and functional action verification \citep{yi2026pivotrl}.
Our learner instead continues from handoff to episode termination
and receives the full-task reward.

\paragraph{Learning signals and teacher supervision.}
RAGEN studies multi-turn RL dynamics and rollout design
\citep{wang2025ragen}, while GiGPO augments episode-level advantages
with comparisons at recurring environment states
\citep{feng2025gigpo}.
Teacher supervision offers another source of learning signal.
On-policy self-distillation conditions teacher targets on privileged
solution information \citep{zhao2026opsd}; Guided-OPD interleaves
teacher and student turns while reducing intervention
\citep{li2026guidedopd}; and HINT-SD targets failure-relevant actions
using hindsight \citep{yeo2026hintsd}.
These approaches motivate distinguishing where the learner starts
from how its continuation is supervised.

\section{Additional Experimental Details}
\label{app:protocol}

\subsection{Data Splits and Checkpoint Selection}
BFCL training uses 100 Base tasks.
The Base evaluation consists of 25 validation tasks and
75 held-out test tasks.
Missing Function, Missing Parameter, and Long Context
are excluded from training.

AppWorld uses 90 training tasks and 57 development tasks.
Test Normal and Test Challenge contain 168 and 417 tasks,
respectively. For AppWorld, we evaluate the final checkpoints after 129.6K generated rollouts for all compared training methods.

\subsection{Optimization and Rollout Generation}
We use AdamW
\citep{loshchilov2019decoupledweightdecayregularization}
with a learning rate of $10^{-6}$ and weight decay of $0.01$.
Each rollout group contains 16 student-generated continuations from the same handoff.
Training uses a sampling temperature of $1.0$ and limits
each model response to 4,096 generated tokens.
We use BF16 precision and implement training with the molt
framework \citep{hu2026molt}, with vLLM
\citep{kwon2023efficientmemorymanagementlarge}
for rollout generation.
Experiments run on four NVIDIA B200 or four NVIDIA H100 GPUs.

\subsection{Movement Step Sizes}
\label{app:selector}
Turn-Threshold moves one user turn whenever its threshold triggers.
Call-Threshold moves one call, except that an all-pass group
moves directly to the preceding user-turn boundary.
All moves are restricted to replay-eligible positions.
Uniform and Bandit can select any eligible position,
including the current one, without an adjacent-step constraint.

\subsection{Bandit Selection}
Each task-position pair maintains success and failure counts.
Using a Beta$(1,1)$ prior, the selector samples a success
probability $p$ from its Beta posterior and computes
\[
    s(p) = 1 - p^{16} - (1-p)^{16},
\]
the probability that a rollout group contains both successes
and failures.

With probability $0.1$, the selector explores a least-observed
position.
Otherwise, it chooses the earliest position whose sampled
score is within $0.1$ of the highest score.

\section{Handoff Positions During Training}
\label{app:start_position}

\begin{figure}[h]
\centering
\includegraphics[width=\columnwidth]{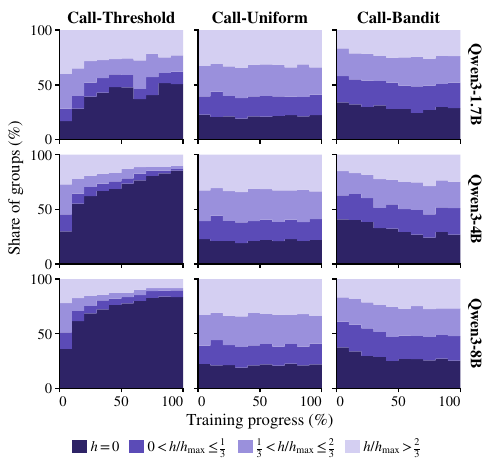}
\caption{Call-level rollout starting positions during BFCL training.}
\label{fig:start_position}
\end{figure}

Figure~\ref{fig:start_position} shows how rollout starting
positions evolve during training.
Call-Threshold increasingly concentrates rollouts at $h=0$,
particularly at 4B and 8B.
This pattern is consistent with its rule of moving earlier
when group success exceeds the promotion threshold.
The shift is less pronounced and more variable at 1.7B.

Call-Uniform maintains a comparatively stable distribution
throughout training, as it samples eligible positions
without conditioning on rollout outcomes.

Call-Bandit continues to allocate rollouts to later starts
rather than progressively concentrating at $h=0$.
Its selection rule favors positions likely to produce mixed
groups, while stochastic selection and exploration allow
continued visits across positions.
As the learner changes during training, success rates at
previously visited handoffs may also change.
Revisiting these positions allows the selector to reassess
where mixed outcomes remain available.
}

\end{document}